# Enabling A Network AI Gym for Autonomous Cyber Agents

Li Li, Jean-Pierre S. El Rami, Adrian Taylor
Defence Research & Development Canada
Ottawa, Canada
{li.li2, jeanpierre.sabbaghelrami, adrian.taylor}@ecn.forces.gc.ca

James Hailing Rao
Dept of Electrical and Computer Engineering
Queen's University
Kingston, Canada
hailing.rao@queensu.ca

Thomas Kunz
Dept. of Systems & Computer Engineering
Carleton University
Ottawa, Canada
tkunz@sce.carleton.ca

*Abstract* — **This work aims to enable autonomous agents for network cyber operations (CyOps) by applying reinforcement and deep reinforcement learning (RL/DRL). The required RL training environment is particularly challenging, as it must balance the need for high fidelity, best achieved through real network emulation, with the need for running large numbers of training episodes, best achieved using simulation. A unified training environment, namely the Cyber Gym for Intelligent Learning (CyGIL), is developed where an emulated CyGIL-E automatically generates a simulated CyGIL-S. From preliminary experimental results, CyGIL-S can train agents in minutes compared with the days required in CyGIL-E. The agents trained in CyGIL-S are transferrable directly to CyGIL-E, showing full decision proficiency in the emulated "real" network. Enabling offline RL, the CyGIL solution presents a promising direction towards sim-to-real in leveraging RL agents in real-world cyber networks.**

## I. Introduction

Recent advancement in reinforcement learning (RL) and deep RL (DRL) brings about the prospect of using AI agents in network cyber operations (CyOps). In CyOps, the attacker, referred to as red, moves through the steps of various actions to form and complete a kill chain [1]. The defender, referred to as blue, must sustain the network mission objectives by throttling the kill chain, removing red's presence and relics, and recovering the compromised functionality. Can blue and red DRL agents automatically learn to achieve a human-level or even superior decision-making proficiency in CyOps?

Indeed, RL/DRL agents have surpassed human experts in many complex and strategic decision-making applications, such as chess, Go, and Atari games [2, 3]. These applications and CyOps share some characteristics, where actors decide and take actions through multiple stages to attain the goals in a dynamic and complex environment. Additionally, RL/DRL methods are being applied in a growing number of real-world systems [4-7]. These developments motivate the new research on the application of RL for autonomous network CyOps [8-13]. The initial use cases may include autonomous red teaming for network hardening, red agents for training human blue teams, and blue agents to assist human blue teams.

RL application requires an agent training environment representative of the real operational environment. A CyOp training environment running on the real network or its emulated version through virtualizing on VMs provides the best realism. However, training in real or emulated networks is too slow [10,18]. This brings about a preferred solution using simulation [18-19], which has the additional benefit of reduced hardware cost.

Current CyOp training environments are mostly simulation-based [8, 10-12]. The challenge for a simulator is in its representation of reality. In the simulator, a finite state machine (FSM) actuates actions by transitioning in the state space. Unlike applications such as board and video games, robotics, or even self-driving vehicles, the CyOp network environment is more complex and unknown. Given CyOp's vast observation and action spaces and emergent action effects on the states, building a correct FSM by human experts is difficult, if not impossible. Existing solutions reduce the complexity by abstracting actions and simplifying their associated states [8-9,11-13]. For example, while real CyOp tools apply many different lateral movement actions, each of which affects a network differently depending on the involved vulnerability and usage conditions, the simulator abstracts them into one action of "lateral movement" [9, 12-13]. Although the agent trained in such a simulator provides high-level insights into attack vectors and countering strategies, it cannot be used directly in the real network, given its very different action and observation space.

Training agents for multistage CyOps grounded in realistic cyber networks shares a problem with many real-world RL applications: a good simulator is essential but hard to build [18]. This problem is even more aggravated for network CyOps. Even though training environments for achieving sim-to-real have been investigated and advanced in other domains such as robotics, for example, by directly using environment images [6], the solutions do not apply to CyOps.

To this end, this work presents a unified CyOp training environment, namely the Cyber Gym for Intelligent Learning

(CyGIL) which consists of an emulator (CyGIL-E) and its high-fidelity simulator (CyGIL-S). The contribution is twofold, in the framework that builds a realistic RL training environment on the real or emulated network, and in the modelling and generation of its mirroring simulator. CyGIL-S trains the agent in minutes instead of training for days and weeks in CyGIL-E. The trained agent can then be directly transferred to the emulated network to carry out the CyOp. To our knowledge, CyGIL is the first reported network CyOp training environment with a unified emulator and simulator capability.

The rest of the paper is organized as follows. Section 2 describes the generic modelling framework for building CyGIL. The unified solution of CyGIL-E and CyGIL-S is presented in Section 3. Section 4 presents some preliminary experimental results. Section 5 elaborates on the next step and directions to conclude the paper.

## II. CyGIL Framework and CyGIL-E

### A. *The Framework and CyGIL-E*

CyGIL-E is built on the framework shown in Fig.1 that maps networked CyOps to the RL training environment, where agents are trained through training games to form its decision engine. A training game involves its objective, the agent reward function (*R*), action space (*A*) and observation space (*O*). Each agent may have its respective *R, A* and *O*. The agent's action space corresponds to its operational tools for command execution. While the blue agent may have plenty of network state data in its observation space, e.g., from deployed sensors and collected logs, alerts and reports, the red agent typically starts from seeing little and grows its observation space step-by-step using the output of its executed commands, i.e., actions. Trained on the action and observation spaces substantiated by red and blue operational tools and their input and output data, agents from CyGIL-E are directly applicable in the operational theatre.

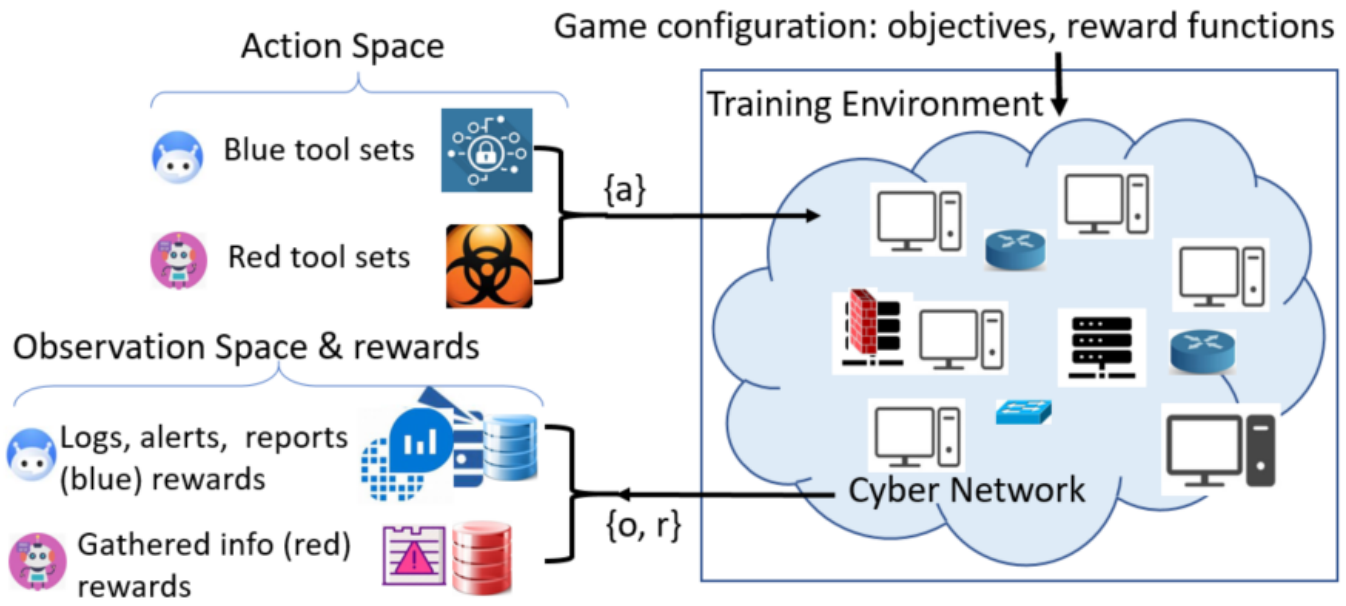

Fig. 1. Map CyOp to its RL training environment: action $\boldsymbol{a} \in \boldsymbol{A}$, observation $\boldsymbol{o} \in \boldsymbol{O}$, reward *r* produced by *R*

For example, MITRE's CALDERA red team emulation platform [14] in CyGIL-E allows the red agent to execute almost all TTPs (Tactics, Techniques and Procedures) contained in the enterprise matrix of the ATT&CK® framework [1], covering all stages of the kill chain after the initial access. The agent's observation includes the data gathered by its actions. The agent may then use the data as input for executing further actions. CyGIL-E wraps the CyOp network with the tools and presents to the agent the training game, including *R, A, O* and the game-ending signal through the standard openAI Gym [10,15] interface, which enables the agent training using all available RL/DRL algorithms.

While SoTA tools such as CALDERA automate the CyOp workflow for human teams, including formatting and launching the operation command, the human expert has to select each command, i.e., the action, to form the course of actions (CoA). CyGIL-E puts these SoTA tools in the hand of AI agents that learn to use the tools autonomously and decide on every action selection for achieving the optimized CoA.

The system design of CyGIL-E is presented in [10] with details of its functional components and interface capabilities. To train the agent, CyGIL-E allows for flexible game design towards different end-to-end CyOp objectives in different network scenarios [10]. This is achieved by emulating the network, defining the CyOp objective through the reward function and selecting the agent's action space. The observation space is then filled by the data available to the agent at each action step.

### B. *Reward Function for Agent Training Games*

In a training game, the tools given to the agent substantiate its action (*A*) and observation spaces (*O*). The agent is trained towards a CyOp objective, i.e., the game training objective in the network. The reward function (R) quantifies and directs the optimization toward the game objective through the agent's maximizing the accumulated reward in the game. Closely related to the game objective, the game-end criteria manage the training episode length to ensure training efficiency and effectiveness. The game-end criteria aim to allow the agent to learn from playing enough action steps to gather maximum rewards while preventing excessive game time that does not contribute to the agent's learning.

The reward function defines the agent's behaviour and influences its training. There are many options for designing the reward function. Reward functions can also vary for different agents. For the examples in this paper, a simple reward function is applied to the red agent:

$$R(o_t, a_t, o_{t+1}) = W(o_{t+1}) - U(s_t, a_t, s_{t+1}) \qquad (1)$$

where $R(o_t, a_t, o_{t+1})$ is the reward given to the red agent when at time *t*, the agent executes action $a_t$ with its observation space $o_t$ to lead to its new observation space $o_{t+1}$ at time *t+1*; $W(o_{t+1})$ is the "worth" of the resulted new observation space $o_{t+1}$; and $U(s_t, a_t, s_{t+1})$ is the cost for the agent to take action $a_t$ at time *t* with the network state as $s_t$ to lead to the state of $s_{t+1}$ at time *t+1*. It should be noted that the network state, such as $s_t$ and $s_{t+1}$, is unknown to the agent. The agent only sees its observation space which is a partial and very limited view of the state, especially in the case of a red agent. In (1), *U* is defined to represent the true cost incurred by the agent, which may not be directly derivable from the agent's observation space. As the network may not give a "reward" to the red agent, in (1), $W(o_{t+1})$ is defined using what the red agent has in its *O* so that the reward and the training are viable for the red agent even without knowing the network state space.

As an example, in some training games, $U(s_t, a_t, s_{t+1}) > 0$ can be set for all *t* to denote a cost for every action. Then if setting $W(o_{t+1}) > 0$ only when the operation objective is reached at *t+1* and $W(o_{t+1}) = 0$ otherwise, the reward function will push the agent to reach the objective, i.e., $W(o_{t+1}) > 0$ as fast as possible to maximize the accumulated rewards. Examples of training game objectives include exfiltration of target files, hijacking the admin account on a server, denying access to a server or a network, etc. The reward function formalizes the training game by modelling the training objectives.

Similarly, game-ending criteria as part of the definition of the training game shape the agent training process. In the experiment presented in this work, the following game-ending criteria are applied. The game ends at time *T+1* if $GE_1 \cup GE_2 = 1$, where $GE_1 = 1$ if the maximum number of action steps allowed is reached, and $GE_2 = 1$ if $\sum_{t=0}^{T} R(o_t, a_t, o_{t+1}) \geq \sum_{t=0}^{\infty} R(o_t, a_t, o_{t+1})$. Thus, the game ends either when the agent has taken the maximum number of actions allowed in the game or when its accumulated reward has reached the maximum value because the agent has collected all the $W(o) > 0$.

### C. An End-to-End Kill Chain Experiment in CyGIL-E

An experiment scenario is depicted in Fig. 2. Hosts 1 and 2 are reachable from the external "Internet" by the Attacker's Command-and-Control (C2). All hosts inside the network can reach the Active Directory Server (ADS) /Domain Controller (DC) at host 6, a Windows 2016 server. Other hosts run on Windows 10 except hosts 1 and 9, which run on Linux. Hosts on the same switch belong to the same subnet and can communicate with each other. Between different subnets, firewall rules are put in place through ONOS to allow host 5 to communicate with hosts 2 and 3 in addition to hosts in its subnet. Each host sends messages to at least one other host at any given time.

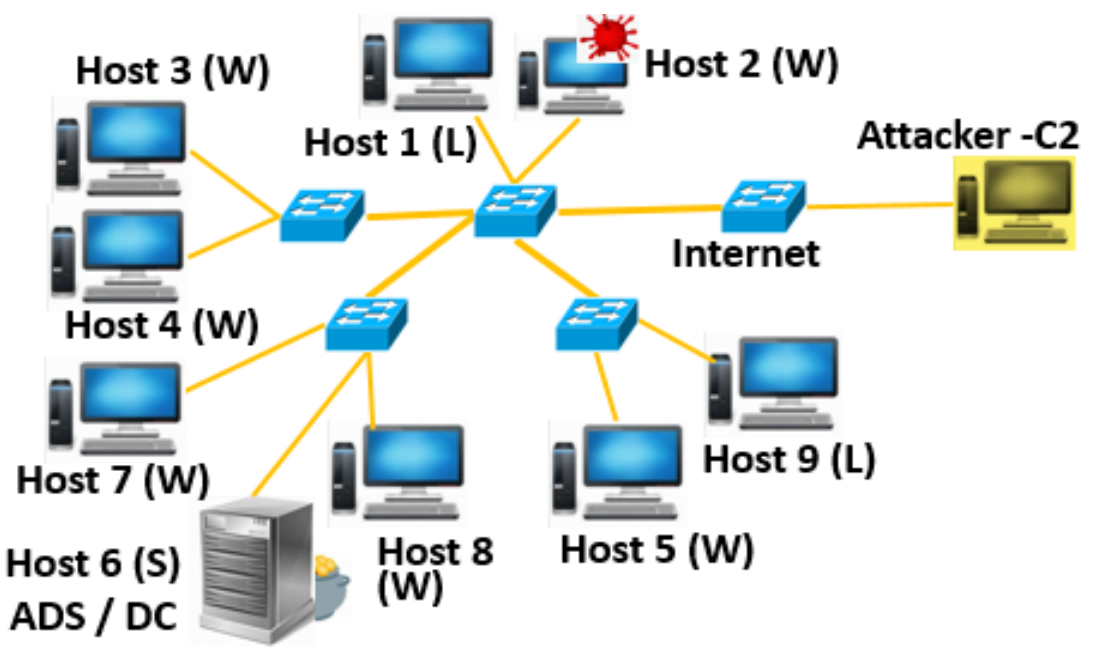

Fig. 2. The example network

In the scenario, host 2 has already been compromised by the attacker using phishing. The implant on host 2, called a "hand" in CyGIL, is thus controlled by the C2 to pivot and execute a kill chain in the network, with the ultimate goal of taking over DC in ADS to compromise the entire domain.

The red agent's action space (Table I) of the training game contains a subset of TTPs from the ATT&CK® framework, which may deliver the kill chain attack by landing on the DC (Fig 2) with the domain admin privilege. The 16 TTPs in the action space amount to several hundred action variants when parameterized for execution. The red agent is trained starting from knowing nothing about either the network or what each action may do with which set of parameters, towards the objective of taking over the DC across the network. The reward function in (1) is applied with $U(s_t, a_t, s_{t+1}) = 1$ for each hand. If the agent lands its hand on the DC with the elevated privilege at time *t+1*, $W(o_{t+1}) = 100$ is obtained by the agent; otherwise, $W(o_{t+1}) = 0$. Therefore, for the game-ending criteria, $GE_2 = 1$ is reached when achieving the game objective of taking over DC. The maximum number of action steps allowed in the game is set up through experimentation, as described in the next section.

TABLE I. AGENT ACTION SPACE TTPS

| ATT&CK Tactics | Actions - ATT&CK Techniques |
|---|---|
| Discovery | T1135: Network Share Discovery |
| Discovery | T1087: Enumerate AD user accounts |
| Discovery | T1018: Remote System discovery |
| Discovery | T1016: Collect ARP details |
| Reconnaissance | T1590: Reverse lookup |
| Credential Access | T1003: Minikats to extract credentials |
| Credential Access | T1110: Brute force credentials |
| Privilege Escalation | T1548. Download & run Sandcat as admin |
| Lateral Movement | T1021: Sandcat remote fileshare WinRM |
| Lateral Movement | T1021: Sandcat remote fileshare (PsExec) |
| Lateral Movement | T1021: Sandcat with SCP (PsExec) |
| Lateral Movement | T1021: Sandcat remote using WinRM |
| Lateral Movement | T1021: Minikatz PSH Sandcat remotely |
| Lateral Movement | T1021: Sandcat remote PsExec |
| Lateral Movement | T1570: Tool transfer by WinRM and SCP |
| Lateral Movement | T1570: Tool transfer by file share |

The experiment is performed on a CyGIL-E "mini version" [10]. The network is emulated on a Windows laptop with Intel(R) Core (TM) i9 and 64GB RAM. Agent training is performed on a laptop with a similar configuration, which connects to the network laptop over an Ethernet [10], mimicking the red agent in the Internet attacking the target network.

As reported in [10], the agent can learn the optimized attack policy using various SoTA DRL models, e.g., DQN (Deep-Q-Network) [16] and PPO (Proximal Policy Optimization) [17], from knowing nothing about either the network or the actions. Among many attack paths and action sequences, the trained agent executes the optimized CoA in every test run amid randomly distributed action outcomes and network conditions. The optimal CoA consists of a minimum of 6 different actions in the required sequence though some actions may need to be repeated due to their random outcomes [10].

TABLE II. AGENT TRAINING TIME IN CYGIL-E

| Network Dimension | Action Space | DQN train. time | PPO train. time |
|---|---|---|---|
| 4 subnets 9 hosts | 16 TTPs | 9 days | 12 days |

The training latencies in CyGIL-E for the DQN and PPO agents are shown in Table II. Although the time is comparable to that needed for human teams to conduct a typical red team exercise, and although the latency can be improved through more hardware for the emulated network, a CyGIL-S that supports fast and high-fidelity training in simulation is required to enable development and experimentation [18-19] of agent training algorithms.

## III. CyGIL-S Auto-Generation

Instead of designing a simulator FSM by a human expert, CyGIL-S is generated automatically from data logged in CyGIL-E, as shown in Fig.3. To directly transfer a trained agent to the emulated or real network, CyGIL-S and CyGIL-E conform to the same action and observation space definition and structure.

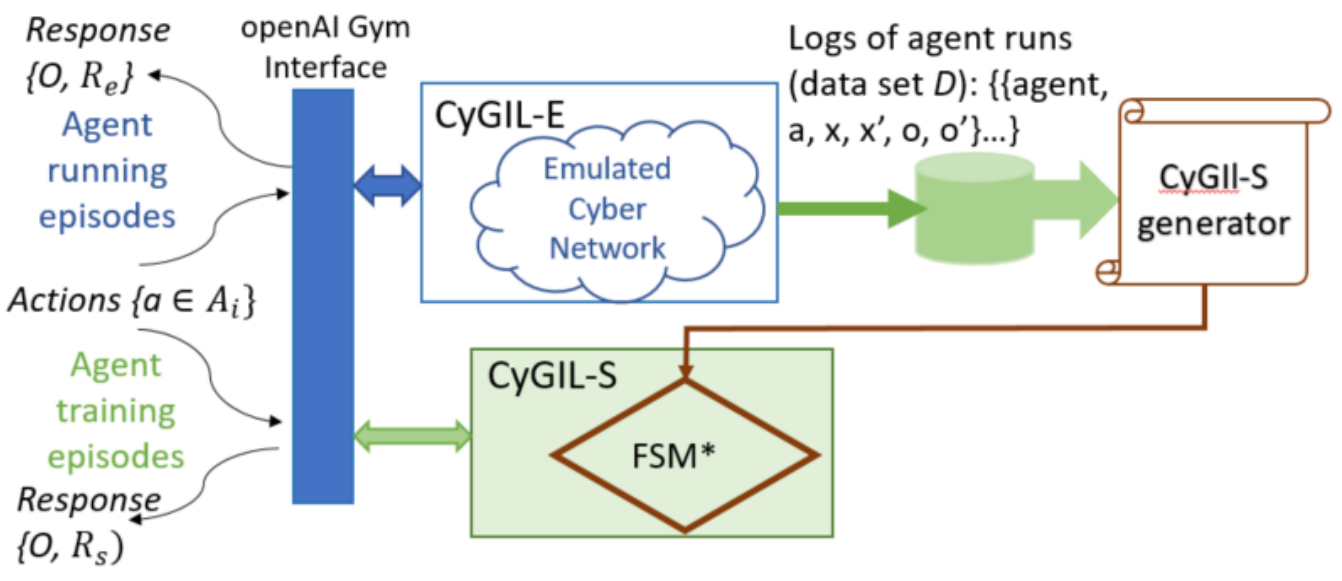

Fig. 3. Generating CyGIL-S from CyGIL-E: $R_e$ and $R_s$ are reward functions in CyGIL-E and CyGIL-S respectively given different games may be run in CyGIL-S; x, x', o and o' are defined in the text of this section.

The auto-generation is built upon the training scenario modelling between CyGIL-E and CyGIL-S. In CyGIL-E, a training scenario is defined by the network and the agent action space, denoted as $GN = \{S, A, P, s_0\}$, where $S$ is the state space, and $A$ is the action space. The probability that action $a$ in state $s$ at time $t$ will lead to state $s'$ at time $t+1, a \in A,\ s, s' \in S$ is $P = P_a(s, s') = \Pr(s_{t+1} = s' | s_t = s, a_t = a)$. The initial state of $S$ is $s_0$.

Adding the reward function, the CyGIL training game is a Markov Decision Process $M = (S, A, P, s_0, R)$, with $R: S_i \times A_i \rightarrow \mathbb{R}$ written as $R(s_t, a_t, s_{t+1}) = R_a(s, s')$, which is the reward received after transitioning from state $s$ to $s'$ by executing action $a$ at time t. The tuple $(A, P, R)$ can be defined per agent. RL trains the agent to learn a policy $\pi\ (a_t|s_t)$ which defines a distribution over actions conditioned on states to maximize the accumulated reward in the game. Multiple training games can be run on each $GN$, differentiated by their reward function $R(s_t, a_t, s_{t+1})$ and the game-ending criteria. It is noted that the state $s$ may not be available to certain agents when they need to calculate the reward. As elaborated in the previous section, the known observation space may be applied instead in such a case.

Each $GN$ defines the FSM of CyGIL-S for a training scenario, i.e., $GN = \{S, A, P, s_0\} = FSM$. For each $GN$, data are collected in CyGIL-E that describe $(S, A, P, s_0)$. $S$ is however unknown but only represented in certain measurement metrics $X$ in CyGIL-E. Even the most comprehensive measurement $X$ is only an approximation of $S$. The data that embeds $FSM^* = \{X, A, P, x_0\}$ is however a realistic approximation of the true FSM, as it captures all that is known of $S$.

Let agent(s) carry out actions on $GN$ in CyGIL-E and gather tuples $(a, x, x') \in \mathcal{D}$, where $x, x' \in X$. Assume that action $a$ taken at the input $x$ generates a total of $N$ different outputs $x_1', x_2', \ldots, x_N'$. $P$ is calculated as

$$P_a(x, x_i') = \frac{C_{(a,x)}^{x_i'}}{\sum_{j=1}^{N} C_{(a,x)}^{x_j'}},\quad \sum_{j=1}^{N} P_a(x, x_j') = 1 \qquad (2)$$

where $P_a(x, x_i') = \Pr(x_{t+1} = x_i' | x_t = x, a_t = a)$, and $C_{(a,x)}^{x_i'}$ counts the output $x_i'$ when action $a$ is taken with the present state measurement $x$.

An agent's observation space $O$ is only a subset of $X$. CyGIL-S then provides the observation transition to the agent according to $P_a(o, o_i')$ which is obtained using (2) with $x = o,\ x_i' = o_i',\ x_j' = o_j'$. $P_a(o, o_i')$ is the probability that action $a$ taken at observation $o$ will lead to $o_i'$. This enables a simple and fast CyGIL-S whose action actuation represents the agent's observation transitions as they occur in CyGIL-E. To form a game on the CyGIL-S of this *GN*, for a red agent, for example, its reward function can be realistically defined by its *O*, as elaborated in the previous section, written here as $R_a(s, s') = R_a(x, x') = R_a(o, o'), o, o' \in O$, given that this is all the agent knows. The game-ending criteria can also be set independently in CyGIL-S using the information in *O*.

The CyGIL-S generator (Fig.3) is therefore data-centric and network CyOp agnostic. The generator code requires no change for new network scenarios and games. From collected data sets of $(a, x, x')$, the CyGIL-S generator computes and constructs the new simulated agent training environment of the training scenario $GN = \{S, A, P, s_0\}$, which consists of the network and the action space *A*. Because $a \in A$ is represented as an index for DRL training, e.g., an integer index, the CyGIL-S generator is applicable even when $A$ is different in the new *GN*. The observation space $O$ is rendered for each agent independent of the semantics of the network configurations and actions, e.g., the network topology, the number of actions in *A*, etc. In summary, the CyGIL-S generator requires no code change for a new *GN* as long as the data structure representing *S (X)*, *A*, and *O* remains the same.

## IV. Preliminary Experimental Results

Both random and DQN training episodes have been collected from CyGIL-E to generate CyGIL-S for the example defined in Fig.2 and Table I. The maximum number of action steps per episode in CyGIL takes various values of 80, 300 and 800. During the data collection, the reward function was altered as well. For example, a high reward for the successful lateral movement compensates for the action's low success rate and leads the agent to discover more paths. A data set containing 157k steps of agent action executions was used in generating CyGIL-S.

TABLE III. Agent Training Time in CyGIL-S using Different Algorithms

| DRL Algorithms | DQN | C51 [19] | PPO |
|---|---|---|---|
| Train. Time | 17.31 min | 5.76 min | 26 min |

### A. *Training Algorithm Experiment in CyGIL-S*

The CyGIL-S agent training runs on a Surface Book3 laptop with Intel® Core™ i7-1065G7 CPU @1.30GHz. From Table III, where the results are averaged over more than 10 training sessions, the agent training time is significantly reduced in CyGIL-S compared with CyGIL-E. Transferred from CyGIL-S to CyGIL-E, all the trained agents execute the optimized CoA in over 50 evaluation episodes.

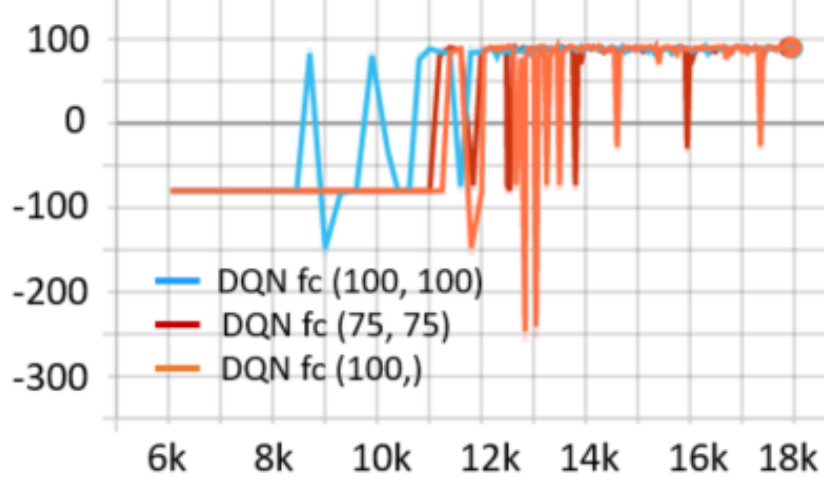

(a) Tuning DQN Algorithms in CyGIL-S: average evaluation return

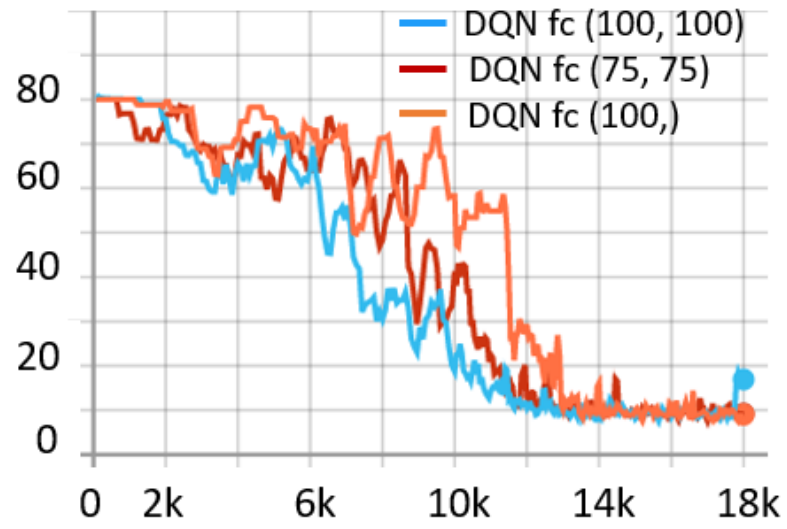

(b) Tuning DQN Algorithms in CyGIL-S: average training episode length

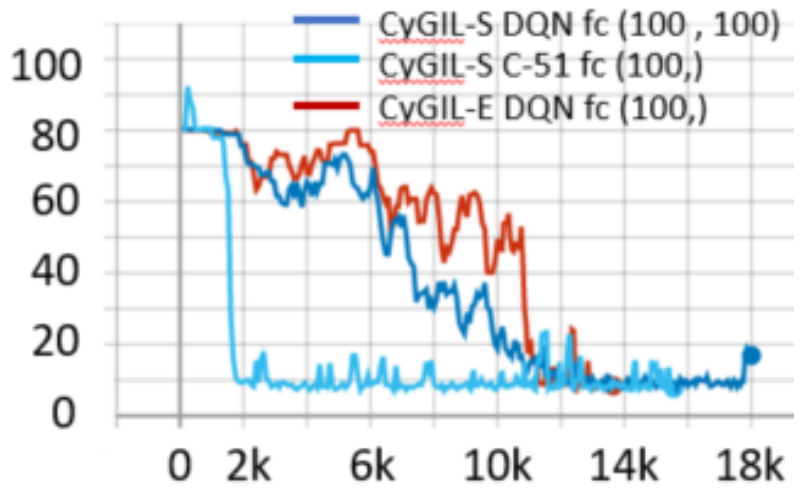

(c) Training different algorithms in CyGIL-S and CyGIL-E: average training episode length

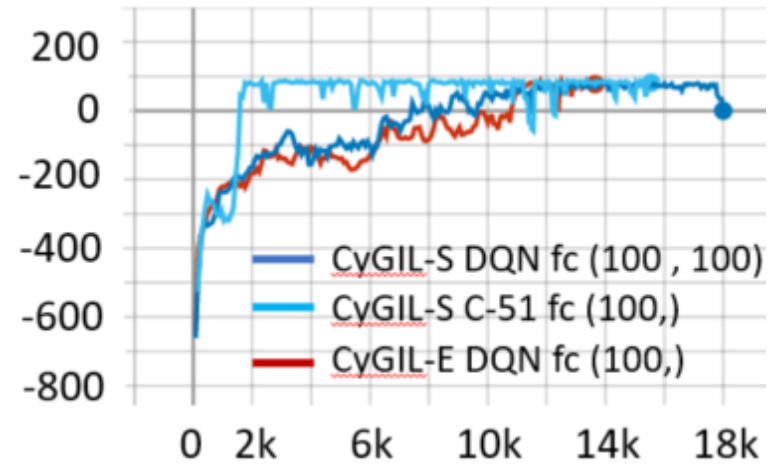

(d) Training different algorithms in CyGIL-S and CyGIL-E: average training return

Fig. 4. Experimental Results in CyGIL-S – x-axis shows training steps; “fc” indicates the architecture of the fully connected layer.

CyGIL-S enables algorithm parameter tuning (Fig.4 (a) and (b)), agent training using different algorithms (Fig.4 (c) and (d)), and game design (Fig. 5). These tasks are infeasible in CyGIL-E due to the extensive latency experienced by CyOp actions in real or emulated networks. In Fig. 4(a) and (b), different DQN architectures are compared. Increasing the complexity of the fully connected (fc) layers, for example, using two layers of $fc = (100, 100)$, expedites the agent policy convergence during the training. As illustrated in Fig 4(b), the average number of action steps in each episode, i.e., the episode length, is reduced faster with $fc = (100, 100)$. It also indicates that this setting may require early stopping to prevent the model from overfitting. At the same time, the overall training and evaluation results are similar when using different parameter settings in DQN. They all reach the optimized policy in a similar amount of training time. The maximum accumulated per-game return received by the agent reaches 92 when the execution results of all actions with random outcomes favour the agent.

While the agent policies trained in CyGIL-S are evaluated and validated in CyGIL-E and demonstrate full decision proficiency, the evaluation in CyGIL-S may sometimes produce erratic return values even when the algorithm is converged, as can be seen in Fig. 4(a). This is given rise by the “noise” in the data collected from CyGIL-E, which is used to generate the CyGIL-S. The noise includes the VM failures and resets when testing CyGIL-E, for example. The noise was not removed from the data set to retain realism and test the DRL algorithms. The algorithms have discerned well such rare events and constructed the correct action decision policy. During the evaluation in CyGIL-S, due to a large number of evaluation episodes used and a relatively small amount of the data embedded in CyGIL-S compared to the total network state space, failure cases are encountered more often than in the real network. This result advises the importance of evaluating the agent policy in CyGIL-E rather than evaluating it only in CyGIL-S.

Results in CyGIL-S inform additional experiments: for example, the agent C-51 (Categorical DQN Rainbow algorithm) [20] was selected for further experiments in CyGIL-E after finding its faster convergence than other algorithms in CyGIL-S (Table II, Fig.4 (c) and (d)).

### *B. Game Design Experiment in CyGIL-S*

The generated CyGIL-S is also employed to design the training game parameters. For example, the maximum number of steps used in the game-ending criteria is evaluated in CyGIL-S, as illustrated in Fig. 5. When using a smaller value than 80 for the maximum number of steps to train the agent, the total training time required is reduced. This is efficient and beneficial, especially for collecting data in CyGIL-E. However, if the maximum number of steps is reduced to 20, the agent policy will not be able to converge. Such a training game has inadequate state space for exploration in the given network training scenario. In addition to game-end criteria, the reward function can also be adjusted in CyGIL-S to improve the agent training efficiency and to potentially train a new agent with a different behaviour and objective without incurring additional latency in CyGIL-E. This requires the data collected from CyGIL-E for generating CyGIL-S to embed the representative distribution of actions related to the new game objective.

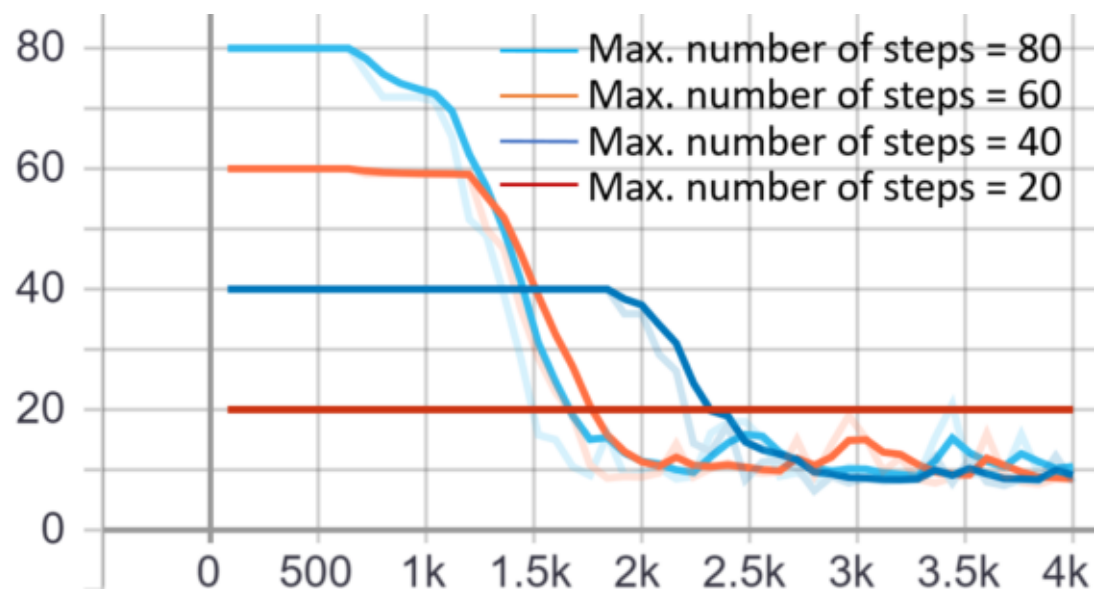

(a) Average training episode length

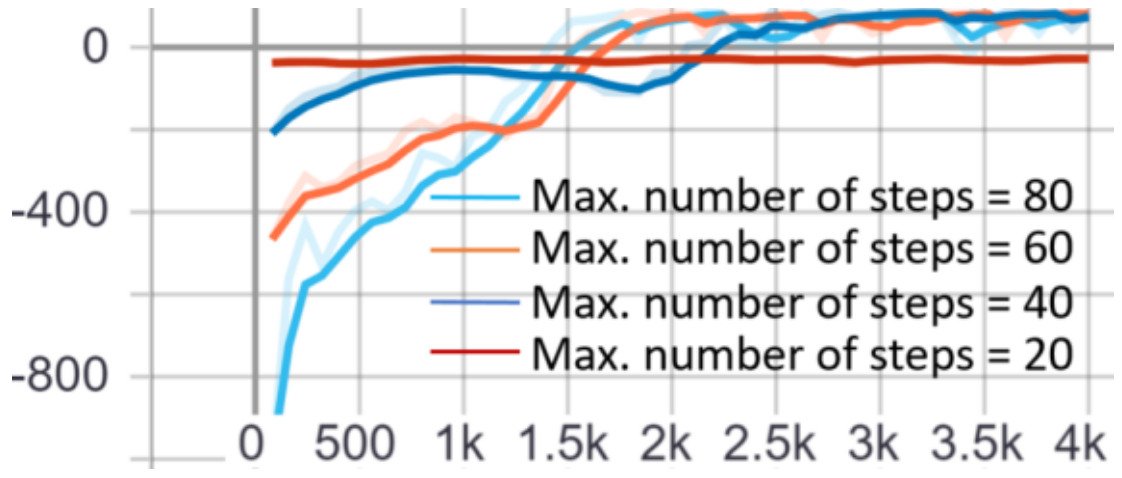

(b) Average training return

Fig. 5. Game design – maximum number of steps per game episode

## V. Further Research and Concluding Remarks

In a preliminary version as presented in this work, the proposed CyGIL solution enables a CyOp training environment that uses simulation with the same high fidelity as the real network or its virtualization with emulation. CyGIL-E runs on real network configurations and CyOp tools. CyGIL-S is automatically generated from CyGIL-E without the need to design a cyber-network simulator as in other SoTA solutions. The CyGIL-S generator is data-centric and network CyOp agnostic. It is written once and reusable in generating CyGIL-S for new scenarios unless the data structure, e.g., table structure for action or observation space in CyGIL-E, changes. CyGIL-S is thus unlike other simulators, which need a rework for every new action and new vulnerability.

Due to its direct representation of the real or emulated environment, the agents trained in CyGIL-S are transferrable to CyGIL-E for continuous training, evaluation and deployment. CyGIL-S enables agent game experiments, training algorithm selection, tuning and new model development.

Although the generated CyGIL-S is sufficient to train agents in the current test cases, a key next step is to generate CyGIL-S capable of supporting more training objectives using the minimum data required from CyGIL-E. To this end, we have started integrated CyGIL-E and CyGIL-S generation and cross-training, continuous learning, and model generalization experiments. This is required in constructing an agent training and experimentation environment that achieves sim-to-real for autonomous CyOps.

## Acknowledgment

The authors would like to thank Mr. Grant Vandenberghe for his valuable advice and support in developing the CyGIL testbed.